# A Comprehensive Survey on Federated Learning: Concept and Applications

Dhurgham Hassan Mahlool [1], *Mohammed Hamzah Abed [1]

[1] AL-Qadisiyah University, College of Computer Science and Information Technology, Computer Science Department

[1*] mohammed.abed@qu.edu.iq
[2] dhurgham.mahlool@qu.edu.iq

\* Corresponding Author

**Abstract.** This paper provides a comprehensive study of Federated Learning (FL) with an emphasis on components, challenges, applications and FL environment. FL can be applicable in multiple fields and domains in real-life models. in the medical system, the privacy of patients records and their medical condition is critical data, therefore collaborative learning or federated learning comes into the picture.on other hand build an intelligent system assist the medical staff without sharing the data lead into the FL concept and one of the applications that are used is a brain tumor diagnosis intelligent system based on AI methods that can efficiently work in a collaborative environment.this paper will introduce some of the applications and related work in the medical field and work under the FL concept then summarize them to introduce the main limitations of their work..

## 1 Introduction

Artificial intelligence (AI) is a branch of computer science that aims to make computers smarter[1] [2]. Learning is one of the most basic requirements for every intelligence behavior. Its techniques work benefited medical specialists by enhancing the power of imaging instruments[3]. For example, AI-assisted picture acquisition can considerably aid in the automation of the COVID-19 scanning method[4]. In addition, AI can boost productivity by accurately delineating infections in X-ray and CT images [5], allowing for easier measurement. In addition, computer-assisted platforms assist radiologists in making clinical judgments[6].

### 1.1 Overview

Federated Learning (FL) is a recent approach that has piqued the interest of researchers who want to learn more about its potential and application[7]. Due to strong data privacy requirements, it is normally regarded impractical to collect and distribute customer data in a centralized location[8]. As well as its works best when on-device data is much more relevant from data on servers, is highly critical to privacy, and is otherwise unwanted or not possible to transfer to servers[9]. Data acquired by mobile devices is typically transmitted and processed centrally together in a cloud server approach. Following that, the data is used to generate insights or excellent inference models[10]. This method, however, is no longer viable because data owners are becoming increasingly concerned about their privacy. A cloud server approach, on the other hand, results in significant propagation delays as well as unacceptable latency. As a reason, the development of new techniques, Therefore, the (FL) technique was used[11].

FL may be used in a variety of fields, but deploying it to different businesses comes with its own set of challenges. Collaborative learning refers to the FL concept, in which algorithms are trained via decentralized samples data across various devices or servers without being to share the real data[12]. This method differs significantly from other well-established methods, like uploading data samples to servers or storing them in a distributed architecture. from the other side, creates more strong modeling without sharing data, resulting in solutions that are more secure and have more access privileges to data[13]. The main challenge is how to train the data without moving this data to a central location? The answer is through (FL), the focus of the FL architecture is on collaboration, which is not always possible with typical machine learning techniques. Furthermore, FL permits the algorithms to accumulate experiences, In other machine learning methods(ML) these cannot always be guaranteed[14]. Rather than aggregating data from multiple sources or depending mostly on traditional discovery and replication methods, It allows training the shared global using a central server while maintaining that data in local institutions in which they originate[15].

The design of a server with FL must be able to deal with its communities varying in size between tens of devices and tens of millions and handle rounds with several devices to thousands or even millions of participants[16]. In addition, the updates collected and delivered throughout each round can be anywhere from a few kilobytes to tens of megabytes in size[17]. Finally, depending on it when devices are inactive and charging, the volume of traffic entering or leaving a given geographic region might vary drastically during the day[18]. Federated Learning works best when on-device data is much more relevant from data on servers, is highly critical to privacy, and is otherwise unwanted or not possible to transfer to servers[19].

FL has been utilized in a variety of applications, including medical, Internet of Things, Transportation, and Defense, and Education, and app of mobile[20]. An FL is highly trustworthy due to its applicability; various experiments have already been undertaken[21]. Despite FL's huge potential, Certain technological components, such as platforms, and software, and hardware, and a slew of others connected to data security and access, remain poorly understood [22]. Because more and more data is becoming freely available from several sources, including healthcare institutions, and patient individuals, insurance firms, and curative industries, there has been a boom of interest for healthcare data processing in recent years[15].figure 1 shows the basic structure of Federated Learning.

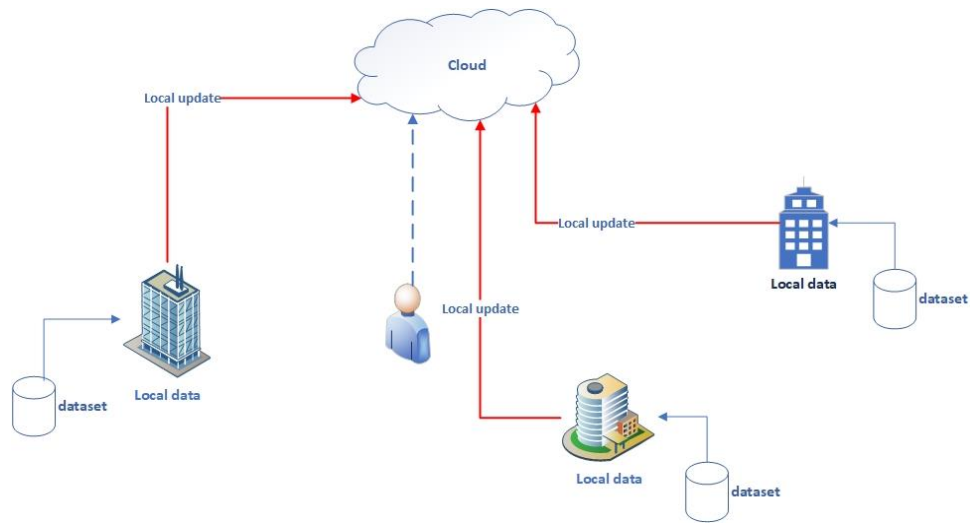

**Fig. 1.** Federated Learning Structure

### 1.2     Problem Statment

One of the most important issues in FL, it's suffered from establishing the initial connection channel and mechanism of communication over the distributed framework, in addition, to sharing the model on a set of client nodes without sharing the client data. In this work, will focus on the expand on the technical aspects with detailed examples of FL and provide extensive assessments, analyses, and comparisons. Communication costs, protocols, software, and applications are among the challenges. We also compare the implementation of FL with earlier works.

### 1.3     Paper layout

The paper is organized as follows. Section I introductions of FL that contain overview and problems. The detail of related works of FL and DL that it's presented in Section II. in section III explains the main challenges facing FL. Finally, section IV explains many applications used in FL.

## 2     Related Work

This section will be discussed in the literature survey focusing on federated learning and deep learning that studies of medical image analysis ,FL is frequently compared with distributed learning, parallelism learning, as well as deep learning. While FL is still a concept young domain, there have been many related publications that look into it in greater depth. Our related work classifies into two approaches: Federated learning and Deep learning.

## 2.1  Federated Learning Approach

Many recent works will be compared in this part, furthermore. In [23], Adnan Qayyum and his Research Team proposed a cooperative learning model for COVID-19 diagnosis by benefiting from a Cluster federated learning method these enabled remote healthcare centers to advantage from each data in other places remotely without sharing the data. Despite substantial developments in modern years, applications in cloud-based medical treatment hold to underutilized because their restrictions run hard privacy and protection (for example low latency), and remote healthcare points, lacking Diagnostic Advanced Facilities.

Because visual data (for example "X-rays" and "ultrasounds") is assembled in several centers, CFL is suitable for the task of detecting COVID-19 in "X-ray" and "Ultrasound" images. in difference in the data distribution, Result Clustered federated learning ensures the best performance than conventional FL, low latency, and increased privacy and data safety. Evaluation of the achievement of the proposed invariance experimental on two datasets, better performance is acquired in two datasets in comparable results against a central baseline, models with central data are trained and refinement 16%, 11% in total F1- Scoring achievement on a multimodal trained in federated learning traditional setting in data sets X-ray and ultrasound[23]. Machine-learning design in public and deep learning in specific need high rate of computational and resources handling, work the publishing of machine learning infeasible of many computing applications. The majority of deep learning models are high and expensive[23].

In[24], the researchers used "SU-Net" which is a federated learning model for brain tumor segmentation that is based on the U-Net architecture. "SU-Net" combines the advantage inception module and a dense block to extract features multi-scale and reuse information from prior layers, enhancing improving information transfer and gradient flows.The proposed model achieved an accuracy of 99.7% which are It was noticed more than a semantic segmentation DeepLabv3+ model and the classical model U-Net allocated to semantic segmentation health images[24].

In [25], Felix Sattler and his research team used "Sparse Ternary Compression" (STC), a modern compression system intended expressly to suit the request of Federated Learning surrounding. The expensive of privacy-preserving collaborative learning is a large increase in communication overhead through training, can decrease lot-required communication by several compression methods, however in Federated Learning these methods are bounded use, by compressing communication upstream from clients to the server only, or performing only well under idealistic conditions example iid distribution in the data client, usually are rarely presented in FL.

The proposed method is a communication protocol that uses sparsification, error accumulation, ternarization, and optimal Golomb encoding to compress each upstream and downstream. "STC" converges faster from Federated Averaging in concerning the number of training repetitions and the number of connected bits, until if the clients keep iid data and use temperate batch sizes in training. "STC" can be understand as an alternative model for efficient federated learning optimization which depends on low-volume high-frequent instead of high volume, low-frequent communication[25].

In [26], Kang Wei and his research team used a new framework based on "differential privacy" (DP), this method was used by adding artificial noises to parameters in the client-direction only prior to aggregation, i.e. "noising before model aggregation federated learning" (NbAFL). First, appear that the NbAFL can require to meet DP at different security levels by correctly adapting differences in added noises. Next, optimize the theoretical convergence related for lack function in the trained federated learning within (NbAFL), and the proposed method used a "K-client random scheduling strategy", it randomly chooses from N all clients to share for any aggregation. Federated learning is a type of machine learning distributed, its ability to prevent disclose data clients' private data to enemies. However, by evaluating uploaded parameters from clients, private information can still be revealed[26]. The comprehensive simulation results confirm the validity of the analysis. Therefore, the analytical results are useful for the design of specificity-preserving federated learning structures with different trade-off requirements on affinity performance and specificity levels[26].

In [27], The Researchers Suggested searching for an analytic solution within federated learning models or meta-analysis. In the federated setup, the model is controlled without sharing information across centers, only the parameters model. Instead, a meta-analysis performs a statistical test to combine results from several independent examinations. Different data sets, it's stored in different institutions, that cannot be directly shared because of legal concerns and privacy, As a result, the full exploitation of large data in the research of brain illnesses is limited. As a result, a fully consistent framework in the federated analysis for distributed biological data was tested and validated[27].

In [28], Researchers used BrainTorrent, a new federated learning framework without a central server, aimed specifically at medical applications. BrainTorrent offers a highly active peer-to-peer environment, each center interacts directly with each other without relying on a central body to organize the training process. A disadvantage of federated learning is the reliance on a central server, are required each client to agree on a single trusted central body, the failure of which may disrupt the training process for all clients.

was select the challenging task for whole-brain segmentation to MRI T1 scan. For experimentation, utilize the Multi-Atlas Labelling Challenge (MALC) dataset. The dataset contains Thirty annotated whole-brain T1 MRI images scan from various patients, with 20 scans being used for training and the remaining 10 being used for testing. QuickNAT architecture was chosen because it provided state-of-the-art performance for whole-brain segmentation. was reduced to a 20-class segmentation issue by combining the right and left brain structures in a single class and all cortical parcellations together in a single cortex class, set the number of epochs to 2 to avoid any client-specific overwriting. All clients started with a learning rate equal to 0:001, which was lowered by a factor of (0:5) after every four update rounds. Distributed these 20 training scans between five clients, with scans for a specific, non-overlapping age range for each client as unequal distribution. The aggregated model for BrainTorrent obtains the performance for the pooled model, but the performance of FLS has 3% Dice points lower than that. When average Dice scores across clients are compared, BrainTorrent surpasses FLS by 7% Dice points[28]. As a result, a proof-of-concept that addresses the challenging job of whole-brain segmentation has been presented, also prove BrainTorrent performance is unaffected by non-uniform distribution of data, while FLS performance suffers[28].

In[29], The Researchers suggested a payoff-sharing method called the "Federated Learning Incentivizer" (FLI). The approach dynamically spilled a given budget among data owners in a federation in a situation-aware way by together developing collective benefited and reducing not equal between data client owners with related in payment received and payout waiting time. need to right incentive suitable mechanism to the context of federated learning is required to maintain long constancy and Adding good quality data owners attracts them during the time. The proposed method can provide almost perfect communal benefit and reduce the owner's data remorse. Due to the constraints of FL, for the short-term incompatibility among contributions and rewards, allowing a robust federated learning environment to form during the time. FLL takes into account all of the aspects that are crucial for federated learning, with a fine distinction of fears related to the time it will begin generating revenue in the federated model [29].

**Table 1. Summary of related work ( Federated learnig )**

| Reference No | Author (s) | Methods |
| --- | --- | --- |
| [23] | Adnan Qayyum at el | clustered federated learning (CFL) |
| [24] | Liping Yi at el | SU-Net for brain tumor segmentation |
| [25] | Felix Sattler at el | Sparse Ternary Compression (STC) |
| [26] | Kang Wei at el | differential privacy (DP) |
| [27] | Santiago Silva at el | meta-analyzing |
| [28] | Abhijit Guha Roy at el | BrainTorrent ( peer-to-peer) |
| [29] | Han Yu at el | Federated Learning Incentivizer (FLI) |

## 2.2 Deep Learning Approach

In this section, Deep Learning and how it compares to FL will explain. Due to "Deep Neural Networks" (DNN) have been utilized for a variety of objectives and generally producing good results, as well as it's frequently compared to DL ap-

proaches[30][31]. DNN has been utilized for a variety of purposes and generally produces promising results, FL has frequently compared with Deep Learning approaches [8].

In[32], the researchers proposed DenseNet-41 was offered as the fundamental framework for the custom CornerNet method. First, the questionable samples in deep features are calculated using DenseNet - 41 module feature extraction in improved CornerNet, in addition, localized and classified the count main points are by using CornerNet's one-step detector. Because of the wide range in texture structure, size, and location of brain tumors, classification, and accurate detection is a difficult task.

CornerNet approach for main points extraction with DenseNet-41, which improved brain tumor classification precision and decrease each training and testing time complication, as well as a low expensive method to brain tumor classification because CornerNet uses a one-step detection structure[32].

In [32], used two datasets: Figshare and the Brain MRI dataset, the Figshare dataset was used, which is a larger and more difficult dataset to identify brain tumors. The Figshare database comprises 930 images from of the Pituitary tumor class, 708 images from the Meningioma tumor class, and 1426 samples from the Glioma tumor type, the matrix size of all images in this dataset is 512 * 512 pixels, The Brain MRI database, the second dataset used, is taken and has a smaller number of samples. There are 155 tumorous samples in this dataset, which compose of 231 MRI pictures with the size (845 * 845), databases its presence via the internet, They contain challenging images in terms of textural complexity, noise, color variations, etc. In the situation of the Figshare, gain an average accuracy of 0.988%, however in situation the MRI, gain an average accuracy of 0.985%, this appears the strength of this method of brain tumor classification. This demonstrates the sturdiness of the method CornerNet for detecting brain tumors. The result Even in the presence of blurring, noise, light variations, the CornerNet architecture, and DenseNet-41 accurately differentiate the tumorous area from the healthy area for the brain. also, by processing challenges in the variable tumor site, structure, and sizes, the proposed design correctly determines cancerous cells of the brain. The solution is capable of accurately distinguishing between different types of brain tumors.

In [33], Javaria Amin and his research team suggested Deep learning models be published to predict segments input as non-tumor (healthy)/ neoplastic (unhealthy), The suggested method consists of three distinct steps. Both high pass filter and median filter are chosen in the initial step to enhance the input image. In the second stage, the brain tumor is segmented using the seed growth method. Finally, in the third stage, the segmented images are prepared into the SSAE model, which employs two hidden layers for classification, detecting a brain tumor is a difficult task because of its different size, shape, and appearance.

BRATS challenge datasets of 2012 (challenge and synthetic), 2013 challenge, 2013 Leaderboard, 2014 challenge, and 2015 challenge are used to assess the model. The 2012 dataset consists of 25 HIGH-GRADE GLIOMA and 25 LOW-GRADE GLIOMA volumes to training and 25 volumes in testing, with each vol-ume containing 155 non-tumor and tumor cuts. Challenge and leaderboard are two subdivisions of BRATS 2013. In the 2013 challenge, 10 LOW-GRADE GLIOMA cases and 20 HIGH-GRADE GLIOMA cases were employed. 4 LOW-GRADE GLIOMA instances and 21 HIGH-GRADE GLIOMA cases make up the leaderboard subset. BRATS 2014 has a total of 300 instances. There are 384 instances in BRATS 2015.The average accuracy in 2012 was 100 % , 90 % in synthetic 2012, 95 % in 2013, 100 % on Leaderboard 2013, 97 % in 2014, and 95 % in 2015. The overall experimental results concluded that the proposed design is better compared with the current approaches[33].

In [34], the Researchers proposed "Grab cut" method is used to segmentation with accurate, and the Transfer learning model (VGG-19) "visual geometry group" is fine-tuned for acquiring features that are then concatenated to hand-crafted (texture and shape) features using a serial-based method. For accurate and rapid classification, these characteristics are optimized using entropy, and classifiers are given a fused vector. brain Tumors have an effect on healthy tissues or increase intracranial pressure, rapid tumor cell development may result in mortality. As a result, it can keep the patient's life if there is early detection of brain tumors.

 the main contributions of this proposed are: transform the input images to one channel. The "Grab cut" technique is used and morphological processes to correctly segment and optimize the area of a tumor. In step a classification, use features (VGG 19) deep and hand-made example gradient orientation graph (HOG) and local binary pattern (LBP) are extracted and combined into a single features vector to distinguish normal images and gliomas[34].

The used model its experiments on better medical image and computer-assisted intervention databases, its compose modals segmentation for a brain tumor that include 2015 and 2016 and 2017.  the best perform offer Accuracy ( 0.9878), and DSC ( 0.9636) In 2015. In 2016 Accuracy (0.9963) , DSC ( 0.9959) ,in 2017 Accuracy (0.9967) and DSC (0.9980). These LBP, HOG, for classification the deep features (VGG-19), and fusions of "HOG", "LBP", and deep features are entered into various classifiers. As a result, it shows that fused features vector is superior performed in comparison with hand crafted  (LBP, HOG) and individual deep features[34].

In [35], the researchers suggested that based Medical Image Optimal Feature Selection Classify utilizing the DL model  By incorporating preprocessing, and feature selection, classification.  This main goal is to derive the best feature selection model for effective medical image categorization. Used a set of  "Gray-level Run Length Matrix" (GLRLM) and "Gray-level cooccurrence matrix"(GLCM). The better subset features are obtained by pre-processed medical images when applying texture features such as GLRLM and  GLCM.  An "Opposition-based Crow Search" (OCS) approaches its increase improved to improve the performance of a deep learning classifier. An OCS algorithm chooses better features from pre-production.

The suggested model achieved the best results in terms of accuracy, and specificity, and sensitivity being 95.22%, and 100%,  86.45 %, and for the group to apply images. The result, based on the soft set, a new system of clinical picture classification was developed to achieve improved execution in terms of accuracy, compu-ting speed, and precision [35].Though its optimal result, The method presented here incurred a high cost[35]. In [36], S. Deepak and P.M.Ameer propose the fully automated classification method by employing MRI images to detect brain tumors. The specific goal is a three-kind classification challenge for classifying brain images containing tumors: meningioma, and glioma, pituitary adenoma. These kinds of tumors are proposed to depend on a group of CNN features and a "support vector machine" (SVM) for classifying. An extract features by using CNN for brain MRI pictures. To achieve the result best, the SVM and CNN features are used.

In [36], Meningioma, and glioma, and pituitary tumors were used to classify brain tumors, to work on experiments, use the Figshare dataset. It's an open data set that's been widely utilized to study medical image categorization and retrieval challenges ( dataset Cheng et). A 3064-weighted with "contrast-enhanced" (CE) MRI from (233) patients make up the dataset. classification the slices for two-dimensional and it's a part of to one of three kinds: meningioma, or glioma, or pituitary tumor. The datasets are unbalanced, contain (1426) glioma MRI images, as well as (708) meningioma MRI images, and (930) pituitary tumor MRI images. The experiment was conducted using a fivefold cross-validation method. The dataset is divided into five subsets with nearly equal size at random. The amount of patients within a certain classify for the tumor is roughly similar across subsets. To do fivefold cross-validation, five disjoint subsets were generated. the first subset is a designated test for each round of validation, while the others are designated as a training set. After five rounds of validation in the dataset with the developed model, each MRI image is examined as well categorized. The classifier's total performance is calculated by combining the validation results from each round. The accuracy is 95.82% that is a better performance than the state-of-the-art methods.

the CNN with SVM combination achieved better accuracy from the produced for a stand-alone CNN classifier, CNN with SVM has lesser computations and memory requirements[36].

In [36], there are been cases this strategy is not useful, On the MNIST as well as Fashion MNIST datasets a multi-class categorization problem (Agarap 2017), CNN is capable to achieve a better accuracy from a CNN-SVM.

**Table 2. Summary of related work ( Deep Learning)**

| Reference No | Author (s) | Methods |
| --- | --- | --- |
| [32] | Marriam Nawaz at el | DenseNet-41-based CornerNet |
| [33] | Javaria Amin at el | stacked sparse autoencoder (SSAE) |
| [34] | Tanzila Saba at el | Handcrafted as well as deep features fusion |
| [35] | R. JOSHUA SAMUEL RAJ at el | Optimal Feature Selection |
| [36] | S. Deepak and P. M. | CNN and SVM |

## 3. Challenges of using Federated learning

Several recent studies have looked into important challenges surrounding the deployment of FL. Presented a thorough examination of the architecture of FL algorithms and discussed numerous challenges, issues, and solutions for improving FL efficacy[37]. The main challenges are:

### 1      privacy

Because the training is completed locally on every device in FL, the raw data always does not leave the person's device. Having additional users in a collaborative approach does, however, increase the risk of attacks, which try to infer sensitive data from devices data[38]. When it comes to federated learning, it's common to presume that the number of participants is high, possibly in the thousands or more. It's hard to know whether or not any of the clients are malicious. FL setting prohibits direct leaking during training or use of the model[15]. The number of multi-system deep learning without exchanging patient data was already examined in the field of medical imaging. Later, has been studied empirically privacy-preserving concerns, by utilizing a sparse vector method and Verification model weights sharing strategies for imbalanced data[39].

### 2      Communication in FL

Training data is spread across several clients in a federated learning environment, each having not reliable and generally poor network connections [40]. There appear to be three options for lowering communication costs: Decrease the number of clients, and decrease the update size and the number of updates. May divide present research on communication-efficient FL into four sets based on three main points: model compression, client selection, updates reduction, and peer-to-peer learning [15] .

### 3      Statistical Heterogeneity

statistical heterogeneity caused by data generated via various types of devices can pose several efficiency issues. For example, optimizing/training (ML/DL) hyperparameters becomes challenging, which has a direct impact on productivity. Meta-learning techniques, for example, can be used to handle statistical heterogeneity and allow device-specific modeling[23].

### 4      Applications of FL

FL is often considered to be one of the most cutting-edge technologies accessible today. Many industries and companies are incorporating FL into their business processes and products. And therefore, a large number of FL applications have emerged. This section will go over some of the applications and usage scenarios. These are used for a variety of things, including healthcare[8].

## 1  FMRI Analysis

Data in healthcare is frequently lacking accuracy and generalizable. Model suffer poor performance as a result of the lack of high-quality data, due to worries about reproducibility. Patients also worry about their medical data being used and shared in subsequent health insurance decisions. Health-care professionals are also concerned that if their data is made public, it may be misused. The authors investigate this issue by using FL to analyze data from functional MRI. The data from fMRI is linked to a variety of neurological diseases and disorders. The authors were successful in developing their recommended framework. without the need for data share, the global server is the two key components of the proposed architecture. The proposed framework discovered the benefits of FL and includes potential applications for finding rarer cancers with fewer patients[8].

## 2  Electronic Medical Records

(Electronic Medical Records) (EMRs)  In a healthcare context, EMRs are the most significant component, as they are utilized to predict illness rates and how a patient responds to therapy. Classical machine learning techniques have been used to deal with EMRs, and these classical machine learning processes have proved successful. Patients create EMRs in a variety of healthcare facilities as well as clinics. Because of its sensitivity, classical machine learning processes are ineffective. There are issues about EMR storage, security, and privacy. While FL can help with these challenges, The authors devised a method known as To address this problem, CBFL (Community Based Federated Learning) was created. The authors employed CBFL algorithm to forecast patient death and length of stay at the hospital. "Encoder training, and K-Means Clustering, Community-Based Learning" are the three techniques in the CBFL[8].

## 3  Auto Correction (text suggestion)

To enhance the accuracy of keyboard search recommendations, use FL. This was Gboard, which is a cellphone device's virtual keyboard. prediction, Autocorrection, word completions, and next-word, and other functions are included in Gboard. It has specific constraints that enable FL suited for it, according to the authors, because it is both an application software and a keyboard. Gboard must not only maintain the privacy of its users, but it must also be free of latency, which is critical in a mobile environment. The authors first gathered information by watching how people use Gboard[8].

## 4  FL in Networking

Federated learning has mostly been utilized in the networking field for work schedules as well as resource allocation. It's especially effective in these cases since it allows for decentralization for making the right decision and modularization of tasks for which each node is responsible. To enable high-reliable and low-latency vehicle communications, the researchers use federated learning. To achieve high reliability as well as low latency in regards to the probabilistic queueing delay, the network-wide power reduction issue is proposed[38]. Figure 2 shows the infrastracture of FL in Network environment and local data for each client.

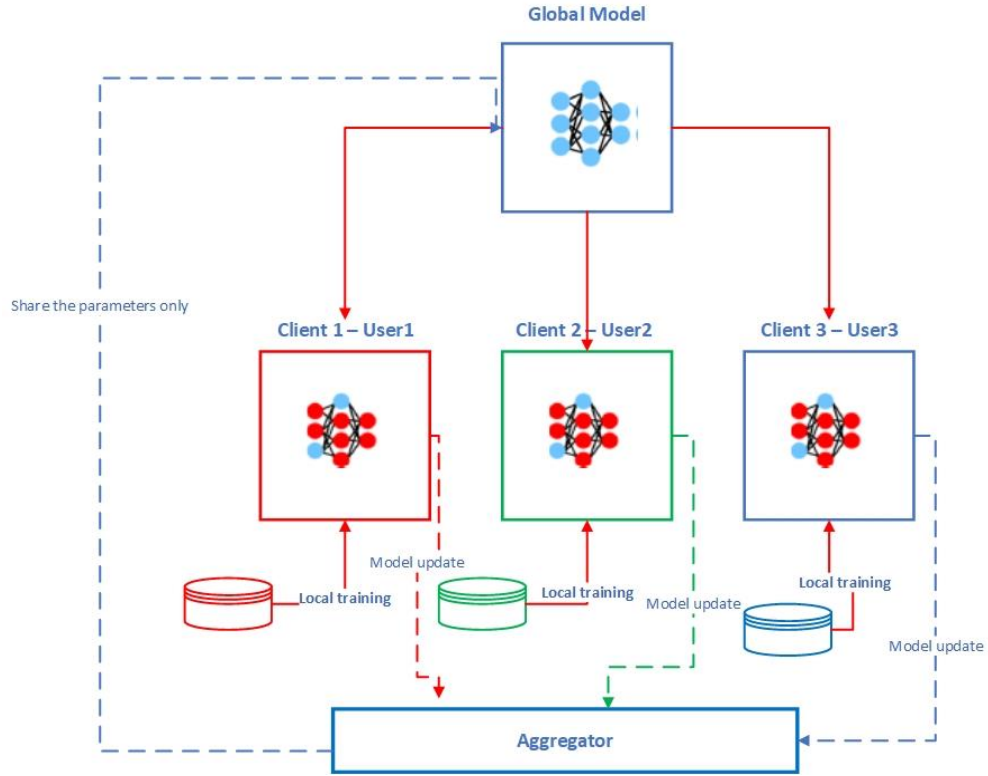

**Fig. 2.** FL bahaviour between clients node and server

## 5   Conclusion

The idea of FL emerging in our modern life quickly aims to improve data privacy and processing for the benefit of a variety of domains. This paper gave a review of the concept's premise, as well as its supporting frameworks, technologies, new research on many aspects of FL. Many applications were also discussed. It thus should provide a solid basic of the many components that make up FL. Additionally, some of the advantages, problems, and related to the design and implementation of FL have been discussed. The scenarios in which FL is most useful are numerous, and obtaining these extra advantages from FL would necessitate ongoing study and optimization before it can be properly simplified. Although the proposed solutions aim to overcome at minimum some of FL's flaws, systems heterogeneity remains a key barrier for FL that must be addressed.

Various FL-based systems get their own set of software and hardware constraints, which can affect performance and emphasize another problem that future frameworks and designs must address: fault tolerance. FL must handle fault tolerance to guarantee that each device that sharing in FL is tracked and that performance that0020is monitored is not jeopardized.